\def\year{2021}\relax
\documentclass[letterpaper]{article} 
\usepackage{aaai21}  
\usepackage{times}  
\usepackage{helvet} 
\usepackage{courier}  
\usepackage[hyphens]{url}  
\usepackage{graphicx} 
\usepackage{natbib}  
\usepackage{caption} 
\nocopyright
\usepackage{cite}
\usepackage{amsmath,amssymb,amsfonts}
\usepackage{textcomp}
\usepackage{xcolor}
\usepackage{microtype}
\usepackage{subfigure}
\usepackage{booktabs}
\usepackage[ruled]{algorithm2e}
\usepackage{algorithmic}

\newcommand{\tabincell}[2]{\begin{tabular}{@{}#1@{}}#2\end{tabular}}

\def\BibTeX{{\rm B\kern-.05em{\sc i\kern-.025em b}\kern-.08em
    T\kern-.1667em\lower.7ex\hbox{E}\kern-.125emX}}

\title{Domain Adaptation with Incomplete Target Domains}
\author{
	Zhenpeng Li\textsuperscript{\rm 1}, Jianan Jiang\textsuperscript{\rm 1},
	Yuhong Guo\textsuperscript{\rm 2},
	Tiantian Tang\textsuperscript{\rm 1}, 
	Chengxiang Zhuo\textsuperscript{\rm 1}, 
	Jieping Ye\textsuperscript{\rm 3} 
	\\
}
\affiliations{
    \textsuperscript{\rm 1}DiDiChuXing\qquad
    \textsuperscript{\rm 2}Carleton University\qquad
    \textsuperscript{\rm 3}University of Michigan 

}

\begin{document}

\maketitle

\begin{abstract}
Domain adaptation, as a task of reducing the annotation cost in a target domain by exploiting
the existing labeled data in an auxiliary source domain, 
has received a lot of attention in the research community.
However, the standard domain adaptation has assumed perfectly observed data
in both domains, while in real world applications the existence of missing data can be prevalent. 
In this paper, we tackle a more challenging domain adaptation scenario where
one has an incomplete target domain with partially observed data.
We propose 
an Incomplete Data Imputation based Adversarial Network (IDIAN) model
to address this new domain adaptation challenge. 
In the proposed model, we design a data imputation module to 
fill the missing feature values based on the partial observations 
in the target domain, while	
aligning the two domains via deep adversarial adaption.
We conduct experiments on both cross-domain benchmark tasks
and a real world adaptation task with imperfect target domains.
The experimental results demonstrate the effectiveness of the proposed method.
\end{abstract}

\section{Introduction}
Although deep learning models have achieved great success in many application domains  
\cite{Krizhevsky2012ImageNet},
their efficacy depends on the
availability of large amounts of labeled training data. 
However, in practice it is often expensive or time-consuming to obtain labeled data. 
Domain adaptation tackles this key issue by
exploiting a label-rich source domain 
to help learn a prediction model for a label-scarce target domain
\cite{ben2007analysis}. 
The standard domain adaptation task assumes perfectly observed data
in both the source and target domains, 
and centers the challenge of domain adaptation on bridging the cross-domain distribution gap. 
However, in real world applications the existence of missing data can be prevalent 
due to the difficulty of collecting complete data features. 
For example, in a service platform, a new user often chooses to fill minimal information
during the registration process while skipping many optional entries. 
The incompleteness of such characteristic data can negatively 
impact the personalized recommendation or advertising strategies adopted by the service platforms.
In such cases, the attempt of using active users' data to help 
make predictions on new users' preferences will not only form a domain adaptation problem
but also entail an incomplete target domain with partially observed instances. 
Directly applying the standard domain adaptation methods in this scenario 
may fail to produce satisfactory results due to the ignorance of 
data incompleteness.

In this paper, we propose 
an adversarial domain adaptation model, named as 
Incomplete Data Imputation based Adversarial Network (IDIAN),
to address the challenge of domain adaptation with incomplete target domains. 
The goal is to learn a good classifier in the target domain by 
effectively exploiting the fully observed and labeled data in the source domain. 
The model is designed to handle both homogeneous and heterogeneous 
cross-domain feature spaces in a semi-supervised setting.
In this model, 
we represent each incomplete instance as 
a pair of an observed instance and a corresponding missing value indication mask,
and use a data generator to fill the missing entries indicated by the mask
based on the observed part. 
To ensure the suitability of the imputed missing data,
we first use domain specific feature extractors to transform both the 
source domain data and the imputed target domain data into a unified feature space,
and then deploy an inter-domain contrastive loss to push the cross-domain instance 
pairs that belong to the same class to have similar feature representations.
To prevent spontaneous cross-domain feature affiliation and overfitting to the discriminative 
class labels,
we introduce a domain specific decoder in each domain to 
regularize the feature extractors under autoencoder frameworks. 
Moreover, we introduce a domain discriminator to adversarially 
align the source and target domains in a further transformed common feature space,
while the classifier can be trained in the same space. 
By simultaneously performing missing data imputation
and bridging the cross-domain divergence gap, 
we expect the proposed model can provide 
an effective knowledge transfer from the source to the target domain 
and induce a good target domain classifier.

To test the proposed model, we conduct experiments on a number of cross-domain benchmark tasks by 
simulating the incomplete target domains. 
In addition, we also test our approach on a real-world ride-hailing service request
prediction problem, which naturally has incomplete data in the target domain.  
The experimental results demonstrate the effectiveness of our proposed model 
by comparing with existing adversarial domain adaptation methods.

\section{Related Work}

\paragraph{Learning with Incomplete Data}
Due to the difficulty of collecting entire feature set values in many application domains, 
learning with incomplete data has been a significant challenge in supervised classification model
training. The work of~\cite{Little2014Statistical} provides a systematic study for data analysis
problems with different data missing mechanisms. 
The naive approach 
of dealing with missing data is using only the partial observations; 
that is, one deletes all entries (or rows) with missing values before deploying
the data for model training. 
Alternatively, the most common strategy is to attempt to impute the missing values. 

Early imputation approaches use some general probabilistic methods to estimate or infer the values of the missing entries. 
For example, the work in \cite{dempster1977maximum} uses the Expectation-Maximum algorithm (EM)
to handle latent variables
and the work in \cite{honaker2011amelia} uses 
multivariate normal likelihoods 
for learning with missing values. 
These methods however require prior knowledge of the underlying model structure. 
Alternatively, the sequential regression in~\cite{raghunathan2015missing} 
provides a variable-by-variable input technique for missing value imputation. 
The multivariate imputation by chained equations (MICE) in~\cite{buuren2010mice} 
provides a multi-category representation of the chain equation. 
Then linear regression was used for the value estimation of 
ordinal variables and multivariate logistic regression was 
used for categorical variables. 
These approaches however can suffer from computational problems when there are 
too many missing variable values. 

Recently, deep learning approaches have been adopted for handling missing values. 
In particular, the generative adversarial networks (GAN) have been adapted 
as a common approach for missing value imputation. 
For example, 
the authors of~\cite{yoon2018gain} proposed 
generative adversarial imputation nets (GAIN), 
which imputes the missing data with a generation network.
The AmbientGAN method developed in~\cite{bora2018ambientgan} 
trains a generative model directly from noisy or incomplete samples. 
The MisGAN in~\cite{li2019misgan} learns a mask distribution to model the 
missingness and uses the masks to generate complete data by filling the missing values 
with a constant value.
Nevertheless, these methods focus on (semi-)supervised learning. 
In another work~\cite{requested} that has recently been brought into our attention, 
the authors propose to address homogeneous unsupervised domain adaptation with data imputation
by learning a data generator after feature mapping in the source domain. 
In contrast, our proposed model addresses supervised domain adaptation with incomplete target domains,
accommodating both homogeneous and heterogeneous cross-domain input feature spaces.

\paragraph{Generative Adversarial Networks}
Generative adversarial networks (GANs)~\cite{goodfellow2014generative} 
generate samples that are indistinguishable from 
the real data by playing a minimax game between 
a generator and a discriminator. 
DCGAN greatly improves the stability of GAN training by improving the architecture of the initial GAN and modifying the network parameters~\cite{mandal2018deep}. 
CGAN generates better quality images by using additional label information 
and is able to control the appearance of the generated images to some extent~\cite{mirza2014conditional}.
Wasserstein GAN uses the Wasserstein distance to increase the standard GAN's training stability~\cite{arjovsky2017wasserstein}.
As already reviewed above, the GAN based models 
have also been developed to address learning with incomplete data.

\paragraph{Domain Adaptation}
Domain adaptation aims to exploit label-rich source domains to
solve the problem of insufficient training data in a target domain~\cite{ben2007analysis}. 
The research effort on domain adaptation has been mostly focused on bridging the cross-domain divergence.
For example, Ghifary et al. 
proposed to use autoencoders in the target domain to obtain domain-invariant features 
\cite{ghifary2016deep}. 
The work in  
\cite{sener2016learning}
proposes using clustering techniques and pseudo-labels to obtain discriminative features. 
Taigman et al. proposed cross-domain image translation methods~\cite{taigman2016unsupervised}.

The authors of~\cite{ben2007analysis} developed theoretical results
on domain adaptation that
suggest the expected prediction risk of a source classifier in the target domain 
is bounded by the divergence of the distributions. 
Motivated by the theoretical work, 
matching distributions of extracted features 
has been considered to be effective in realizing an
accurate adaptation 
\cite{bousmalis2016domain, purushotham2019variational, li2018deep, sun2016return}.
The representative method of distribution matching 
learns a domain adversarial neural network (DANN) by 
extracting features that 
deceive a domain discrimination classifier 
\cite{ganin2016domain}. 
It extends the idea of generative adversarial networks 
into the domain adaptation setting by using the feature extraction network 
as a generator and using the domain classifier as a discriminator. 
The features that can maximumly confuse the discriminator are expected to 
effectively match the feature distributions across the source and target domains. 
The conditional domain adversarial network model (CDAN)
further extends DANN by 
aligning the joint distribution of feature and category across domains
\cite{long2018conditional}.
In addition, some other methods have utilized the maximum mean discrepancy (MMD) 
criterion 
to measure the distribution divergence in high-dimensional space between different domains 
\cite{long2016unsupervised, long2015learning}, 
They train the model to simultaneously minimize both the MMD based cross-domain divergence and 
the prediction loss on the labeled training data.
Nevertheless, all these domain adaptation methods assume fully observed data in 
both the source and target domains. 
In this paper, we address a novel domain adaptation setting where the target domain
contains incomplete data.

\begin{figure*}[th!]
\centering
\includegraphics[width=0.8\linewidth]{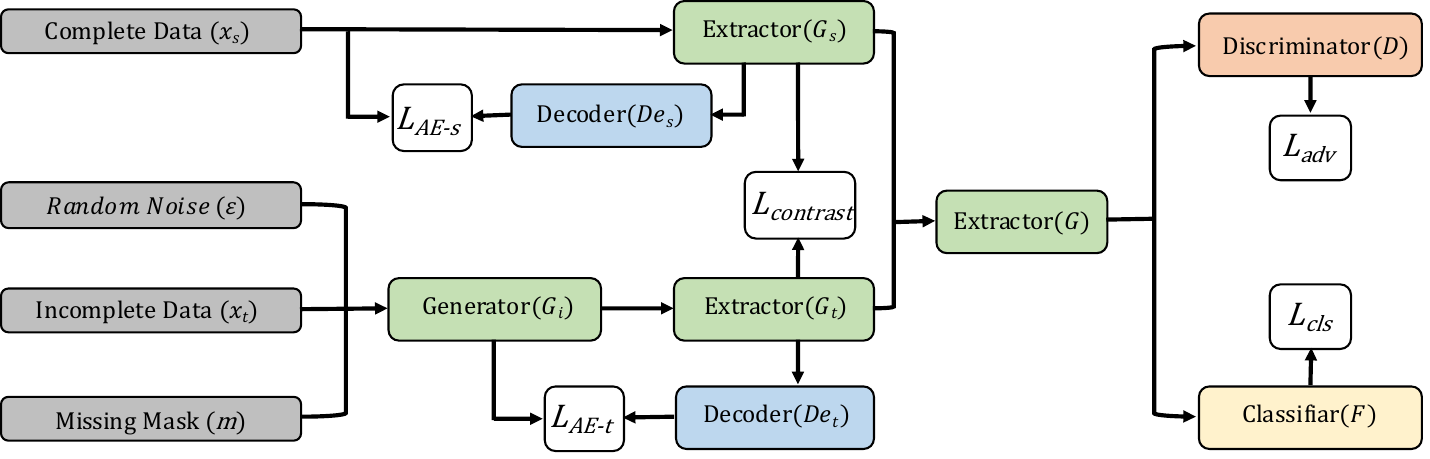}
	\caption{The overall structure of the proposed incomplete data imputation based adversarial network (IDIAN). It has the following components: 
	(1) The incomplete data generator $G_{i}$ in the target domain. 
	(2) The domain specific autoencoders in both domains, each of which is formed 
	by a feature extractor and a decoder (($G_s, De_s$) or ($G_t, De_t$)). 
	(3) The adversarial domain adapter, which is formed by a common feature extractor $G$,
	a domain discriminator $D$, and a classifier $F$. 
	}
\label{fig:impute}
\end{figure*}

\section{Method}

We consider the following domain adaptation setting. 
We have a source domain ${\it S}$ and a target domain ${\it T}$.  
The source domain has a large number ($n_s$) of labeled instances, 
$\mathcal{D}_{S}={\left\{(x_{i}^{s}, y_{i}^{s})\right \}}_{i=1}^{n_{s}}$,
where $x_i^s\in\mathcal{R}^{d_s}$ denotes the $i$-th instance and 
$y_i^s$ is a $\{0,1\}$-valued label indicator vector.
In the target domain, we assume there are a very small number of labeled instances 
and all the other instances are unlabeled: 
$\mathcal{D}_{T}={\left\{ (x_{i}^{t}, m_i^t, \cdot)\right \}}_{i=1}^{n_{t}}$,
where $x_i^t\in\mathcal{R}^{d_t}$ denotes the $i$-th target instance,
which is only partially observed and its entry observation status is encoded
by a binary-valued mask vector  $m_i^t\in\{0,1\}^{d_t}$. 
Without loss of generality, we assume the first $n_l$ instances are labeled, 
such that 
$\mathcal{D}_{T_l}={\left\{(x_{i}^{t}, m_i^t, y_{i}^{t})\right \}}_{i=1}^{n_{l}}$ 
and $\mathcal{D}_{T_l}\subseteq \mathcal{D}_{T}$.
We further assume the class label spaces in the two domains are the same,
while their input feature spaces can be either same ($\mathcal{R}^{d_s}=\mathcal{R}^{d_t}$) 
or different ($\mathcal{R}^{d_s}\not=\mathcal{R}^{d_t}$). 

In this section, we present an incomplete data imputation based adversarial learning network (IDIAN)
to address the challenging domain adaptation problem above. 
The proposed IDIAN model is illustrated in Figure~\ref{fig:impute}.
It has the following components:
(1) The incomplete data generator $G_{i}$, which imputes the missing values in the target domain.
(2) The domain specific autoencoders in both domains, each of which is formed 
by a feature extractor and a decoder (($G_s, De_s$) or ($G_t, De_t$)). 
They map the input data from both domains into a unified feature space
by ensuring both information preservation via a reconstruction autoencoder loss 
($L_{AE_s}$ or $L_{AE_t}$) 
and discriminative cross-domain alignment via an inter-domain contrastive loss ($L_{cont}$).
(3) The adversarial domain adapter, which is formed by a common feature extractor $G$,
a domain discriminator $D$, and a classifier $F$ after the cross-domain feature space unification. 
It performs adversarial cross-domain feature alignment to bridge the cross-domain divergence
and induces a good classifier $F$. 
These components coordinate with each other under the proposed framework 
to facilitate the overall effective knowledge transfer and classifier training. 
Below we present these components and the overall learning objective in detail.

\subsection{Incomplete Data Imputation}
The existence of missing data in the target domain presents a significant challenge
for domain adaptation. Simply ignoring the missing data or imputing the missing entries
with non-informative zeros will unavoidably lead to information loss and degrade the adaptation performance. 
Meanwhile, one fundamental assumption of domain adaptation is that 
the source and target domains share the same prediction problem 
but present different data distributions or representation forms in the input feature space. 
This suggests that the suitable data imputation in the target domain
should coherently support the common prediction model induction
and the mitigation of the cross-domain divergence. 
In light of this understanding, 
we propose to simultaneously perform data imputation in the target domain,
match the cross-domain data distributions and learn the classifier in an unified feature space
under the end-to-end IDIAN learning framework.
In particular, as shown in Figure~\ref{fig:impute},
we introduce a generation network $G_i$ to perform data imputation within the IDIAN. 

Typically different features (attributes) in the input space 
are not independent from each other but rather 
present correlations. 
Hence we propose to generate the 
missing values of each instance based on its observed entries. 
Specifically, our generator $G_i$ takes a triplet ($x^{t}, m^t, \varepsilon$) as input,
where $x^t$ denotes the given partially observed instance in the target domain, 
$m^t$ is the corresponding mask vector with value 1 indicating an observed entry 
and value 0 indicating a missing entry, 
and $\varepsilon$ is a noise vector randomly sampled from a standard normal distribution.
Then $G_i$ generates the imputed instance as follows:
\begin{equation}
G_{i}\left ( x^{t}, m^t, \varepsilon \right ) = 
x^{t}\odot m^t + \hat{G}_{i}\left ( x^{t}\odot m^t + \varepsilon\odot \bar{m}^t \right )\odot \bar{m}^t 
\label{eq:Gi}
\end{equation}
where $\bar{m}^t = 1-m^t$ and ``$\odot$" denotes the Hadamard product operation. 
Here the imputation network $\hat{G}$ fills the missing values of $x^t$,
and the overall computation in Eq.(\ref{eq:Gi}) ensures the original observed
features will not be modified. 

\subsection{Feature Space Unification with Discriminatively Aligned Autoencoders} 

The proposed IDIAN model allows heterogeneous cross-domain input feature spaces. 
Hence we introduce two domain specific feature extractors, 
$G_{s}$ and $G_{t}$, in the source and target domains respectively to 
transform the input features into a unified feature space. 
Moreover, to prevent information loss during the feature transformation
we introduce two domain specific decoders, $De_s$ and $De_t$,
to form autoencoders together with $G_s$ and $G_t$ 
in the source and target domains respectively.
The principle of autoencoder learning lies in minimizing the reconstruction loss between 
the original input instances and their corresponding reconstructed versions 
which are obtained by feeding each instance through the feature extractor (encoder) and decoder.
A small reconstruction error ensures the feature extractor to preserve essential information from the inputs. 
In the proposed model, we use the following reconstruction loss in the two domains: 
\begin{equation}
\begin{aligned}
 L_{AE} &= L_{AE_{s}}+L_{AE_{t}}\\
  &=\frac{1}{n_{s}}\sum\nolimits_{i=1}^{n_{s}} \left \| De_{s}\left ( G_{s}\left ( x_{i}^{s} \right ) \right )- x_{i}^{s}\right \|^{2}+
\\ 
&\quad \frac{1}{n_{t}}\sum\nolimits_{i=1}^{n_{t}} \left \| De_{t}\left ( G_{t}\left ( \hat{x}_{i}^{t} \right ) \right )- \hat{x}_{i}^{t}\right \|^{2} 
\label{eq:Lae}
\end{aligned}
\end{equation}
where $\hat{x}_i^t$ denotes the imputed $i$-th instance in the target domain,
such that $\hat{x}_i^t = G_{i}\left ( x_i^{t}, m_i^t, \varepsilon \right )$. 

\paragraph{Inter-Domain Contrastive Loss} 

As domain adaptation assumes a shared prediction problem 
in the unified feature representation space, 
we further propose to discriminatively align the extracted features 
of the instances from the two domains
based on their corresponding labels,
in order to ensure a {\em unified} feature space after the feature extraction. 
Specifically, we design the following inter-domain contrastive loss
to promote the discriminative alignment of the instances across domains: 
\begin{equation}
	L_{cont}=\mathbb{E}_{(x_{i}, x_{j})\sim \mathcal{D}_{S}\cup \mathcal{D}_{T_l}}
L_{dis}\!\left ( f_{i}, f_{j},\delta \left ( y_{i},y_{j} \right ) \right )
\label{eq:Lc}
\end{equation}
where $\delta(y_i,y_j)$ is an identity indication function, which 
has value 1 when  $y_i=y_j$ and has value 0 when $y_i\not=y_j$; 
$f_i$ and $f_j$ denote the extracted feature vectors for instances $x_i$ and $x_j$ respectively,
such that
\begin{align}
	f_i = \left\{
		\begin{array}{ll}
			G_s(x_i) &\mbox{if}\; x_i\in\mathcal{D}_S,\\	
			G_t(G_i(x_i,m_i,\varepsilon)) &\mbox{if}\; x_i\in\mathcal{D}_{T_l}.\\	
		\end{array}
		\right. 
\end{align}
The contrastive distance function $L_{dis}$ is defined as:
\begin{equation}
	L_{dis}=\left\{\begin{array}{ll}
	\left \| f_{i}-f_{j} \right \|^{2} & \mbox{if}\; \delta\left ( y_{i},y_{j} \right )=1,\\
	\max\left ( 0,\, \rho-\left \| f_{i}-f_{j} \right \|^{2} \right ) & \mbox{if}\; \delta\left ( y_{i},y_{j} \right )=0.
\end{array}\right.
\end{equation}
Here $\rho$ is a pre-defined margin value,
which is used to control the distance margin between instances from different classes. 
This contrastive loss aims to reduce the intra-class distance and increase the inter-class distance 
over data from both the source and target domains in the unified feature space.

\subsection{Adversarial Feature Alignment}
The discriminatively aligned autoencoders above aim to induce a unified feature space. 
However, there might still be distribution divergence across domains.
We therefore deploy an adversarial domain adaptation module to align the 
cross-domain feature distributions, while training a common classifier. 
As shown in Figure~\ref{fig:impute},
the adversarial adaptation module consists of a feature extractor $G$,
a domain discriminator $D$, and a classifier $F$. 
$D$ is a binary probabilistic classifier 
that assigns label 1 to the source domain and label 0 to the target domain.
Following the principle of the adversarial training of neural networks,
the module plays a minimax game between the feature extractor $G$ and the domain discriminator $D$
through the following adversarial loss:
\begin{align}
L_{adv} =  &-\frac{1}{n_{s}}\sum\nolimits_{i=1}^{n_{s}}\log D\left ( G ( f_{i}^{s}) \right )\nonumber\\ 
&-\frac{1}{n_{t}}\sum\nolimits_{j=1}^{n_{t}}\log\left ( 1-D\left ( G (f_{i}^{t}) \right ) \right )
\label{eq:Ladv}
\end{align}
where $f_i^s = G_s(x_i^s)$ and $f_i^t = G_t(G_i(x_i^t,m_i^t,\varepsilon))$.
The domain discriminator $D$ will be trained to maximumly distinguish the
two domains by minimizing this loss,
while $G$ aims to produce suitable features to confuse $D$ 
by maximizing this adversarial loss and hence diminishing the cross-domain distribution gap.
Meanwhile, we also train the classifier $F$ in the extracted feature space
by minimizing the following cross-entropy classification loss on all the labeled instances:
\begin{align}
	L_{cls} = &- \frac{1}{n_{l}}\sum\nolimits_{j=1}^{n_{l}}y_j^{t\top}\log F\left ( G (f_{i}^{t})\right ) \nonumber\\ 
	&-\alpha\frac{1}{n_{s}}\sum\nolimits_{i=1}^{n_{s}}y_i^{s\top}\log F\left (G (f_{i}^{s})\right )
\label{eq:Lcls}
\end{align}
where $\alpha $ is a trade-off hyperparameter. 
\begin{algorithm}[t!]
\caption{Training algorithm for IDIAN}
\label{alg:IDIAN}
\begin{algorithmic}[0]
\STATE{\bf Input:} Training data $\mathcal{D}_S$ and  $\mathcal{D}_T$; 
	trade-off parameters $\alpha$, $\beta$, $\gamma$, and $\lambda$; constant margin $\rho$;
	epoch\# $N_e$, batch size $n_b$\\ 
\STATE{\bf Initialization:} Randomly initialize the model parameters	
\FOR{{ $k=1$ {\bf to} $N_e$}}
\STATE{Randomly reshuffle $\mathcal{D}_S$ into a set of mini-batches
	$S_B = \{B_1, B_2,\cdots, B_J\}$ with batch size $n_b$}
\FOR{{$j=1$ {\bf to } $J$}}  
	\STATE Randomly sample a batch $B_l^t$ from $\mathcal{D}_{T_l}$ with size $n_b$.
	Randomly sample a batch $B^t$ from $\mathcal{D}_T\setminus\mathcal{D}_{T_l}$ with size $n_b$. 
	Set $B_j= B_j\cup B_l^t\cup B^t$
\ENDFOR
\FOR{$B\in S_B$ }
\STATE 1. Generate imputed data for the incomplete target instances in $B$ using $G_i$ with 
	Eq.(\ref{eq:Gi})
\STATE 2. Compute the reconstruction loss $L_{AE}$ on imputed batch $B$ with Eq.(\ref{eq:Lae})
\STATE 3. Compute the contrastive loss $L_{cont}$ on imputed batch $B$ with Eq.(\ref{eq:Lc})
\STATE 4. Compute the adversarial loss $L_{adv}$ and the classification loss $L_{cls}$ 
	with Eq.(\ref{eq:Ladv}) and Eq.(\ref{eq:Lcls}) respectively on imputed batch $B$ 
\STATE 5. Conduct gradient descent over parameters of each component network :
\begin{align*}
	\Theta_{G_i}, \Theta_{G_s}, \Theta_{G_t}\overset{-}{\leftarrow}\; & \eta\; 
	\triangledown_{\Theta_{G_i},\Theta_{G_s},\Theta_{G_t}} \! L(\Theta)
\\	
	\Theta_{De_s}, 	\Theta_{De_t} \overset{-}{\leftarrow}\;& \eta\;
	\triangledown_{\Theta_{De_s},\Theta_{De_t}} \! L_{AE}
\\	
	\Theta_{G}\overset{-}{\leftarrow}\;&\eta\;\triangledown_{\Theta_{G}}\!\left 
	( L_{cls} \!+\! \gamma L_{cont} \!-\! \lambda L_{adv} \right ) \\
	\Theta_{D}\overset{+}{\leftarrow}\;&\eta\;\triangledown_{\Theta_{D}}\left ( \lambda L_{adv} \right ) \\
	\Theta_{F}\overset{-}{\leftarrow}\;&\eta\;\triangledown_{\Theta_{F}}\left ( L_{cls} \right )
\end{align*}
    \ENDFOR
\ENDFOR
\end{algorithmic}
\end{algorithm}


\subsection{Overall Learning Problem}
Finally, by integrating the autoencoders' reconstruction loss, the contrastive loss,
the adversarial loss, and the classification loss together, 
we have the following adversarial learning problem for the proposed IDIAN model:
\begin{align}
	& \min_{G_{i}\!,G_{s}\!,G_{t}\!,G\!,F\!,De_{s}\!,De_{t}}\max_D
\quad L(\Theta)\\
	& \mbox{where}\quad 	
L(\Theta) =L_{cls}+\beta L_{AE}+\gamma  L_{cont}-\lambda L_{adv}
\nonumber
\end{align}
where $\lambda $, $\beta$, $\gamma$ are trade-off parameters 
between different loss terms; $\Theta$ denotes the parameters
for all the component networks (the whole model). 
The model is trained in an end-to-end fashion with stochastic gradient descent. 
The overall algorithm is illustrated in Algorithm~\ref{alg:IDIAN}.


\begin{figure}[t]
\centering
\subfigure[]{
\label{fig:digit_10_mt2mm.sub.1}
\includegraphics[width=0.48\linewidth]{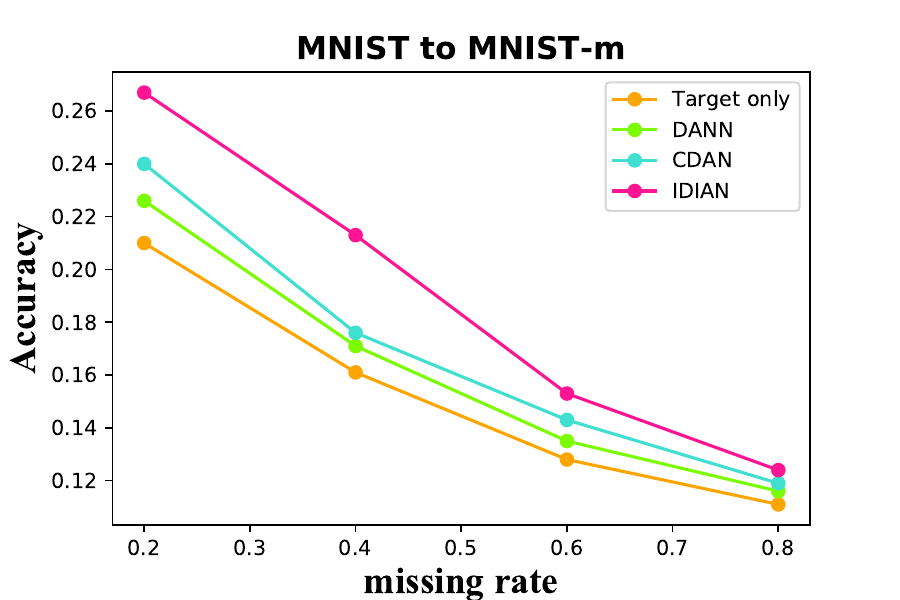}}
\subfigure[]{
\label{fig:digit_10_mt2mm.sub.2}
\includegraphics[width=0.48\linewidth]{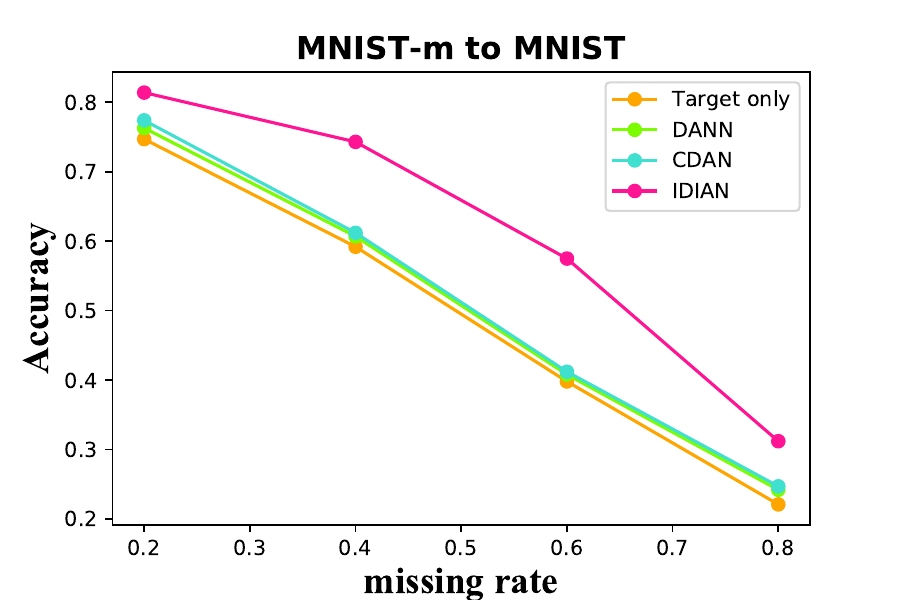}}
\subfigure[]{
\label{fig:digit_10_mt2usps.sub.1}
\includegraphics[width=0.48\linewidth]{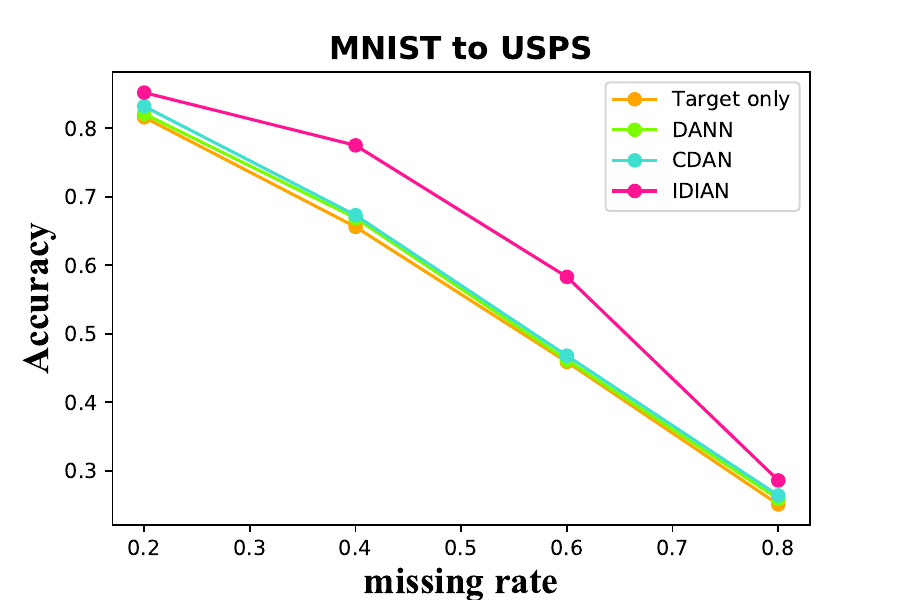}}
\subfigure[]{
\label{fig:digit_10_mt2usps.sub.2}
\includegraphics[width=0.48\linewidth]{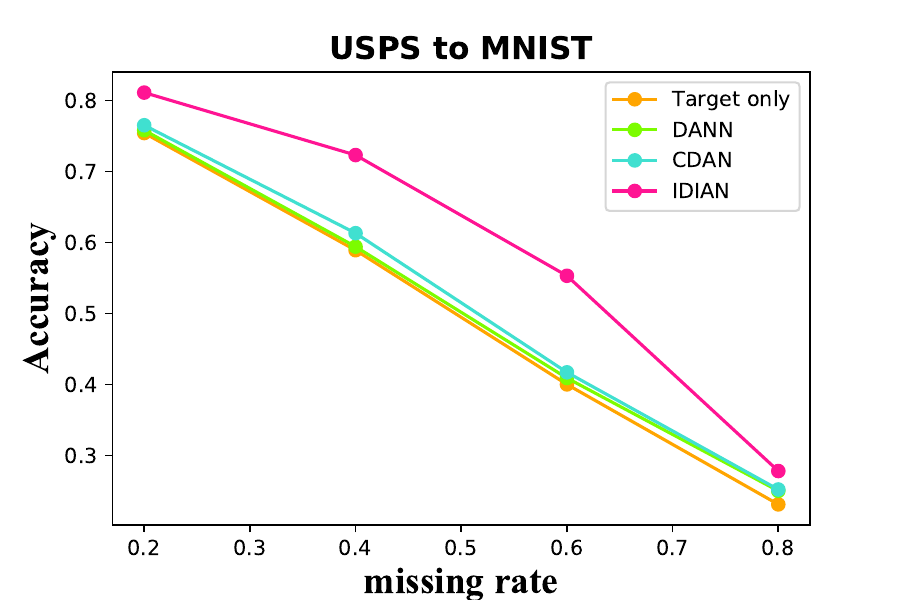}}
\subfigure[]{
\label{fig:digit_10_svhn2syn.sub.1}
\includegraphics[width=0.48\linewidth]{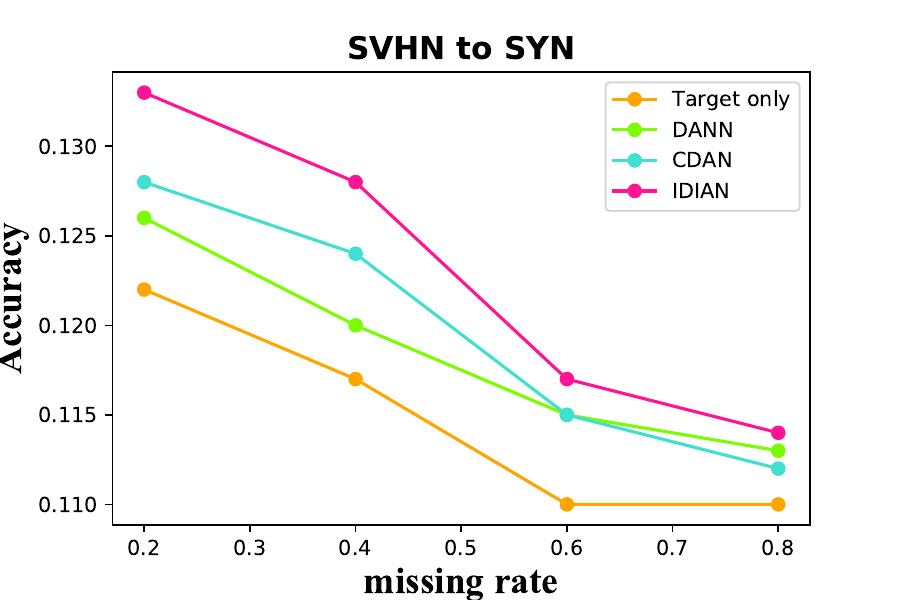}}
\subfigure[]{
\label{fig:digit_10_svhn2syn.sub.2}
\includegraphics[width=0.48\linewidth]{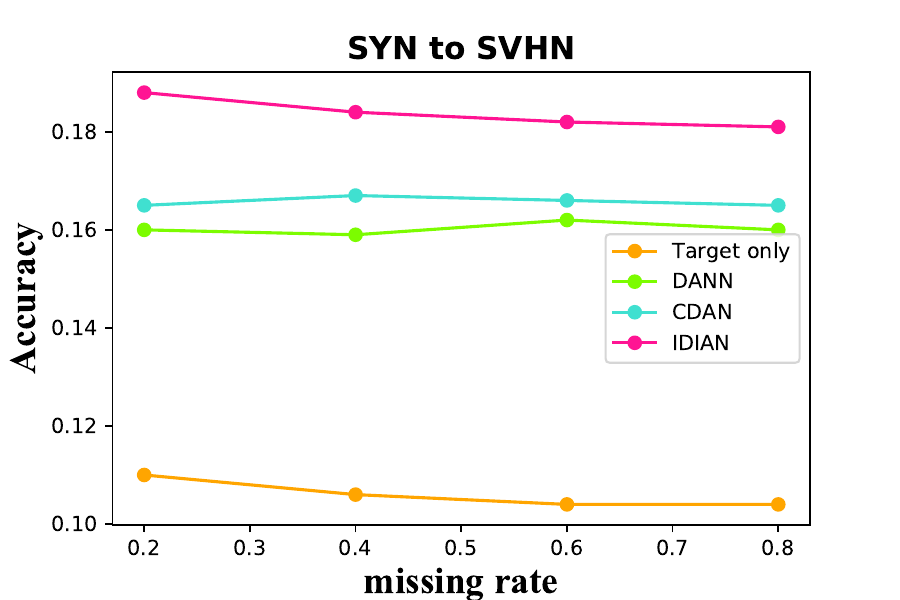}}
\caption{Test accuracy comparison of different domain adaptation methods 
	with different data missing rates in the target domain.  
	10 instances per class in each target domain are used as the labeled data.}
\label{fig:digit10}
\end{figure}

\begin{table}[h!]
\begin{center}
\caption{Architecture details of IDIAN. `fc(x,y)' denotes a fully connected layer
	with input size x and output size y. `relu', `sigmoid' and `softmax'
denote the activation functions. 
$d_s$ and $d_t$ denote the input feature dimension in the source and target domains respectively; 
	$n_c$ denotes the number of classes.}
\label{tab:modelarch}
\resizebox{\linewidth}{!}{
\begin{tabular}{l|c||l|c}
\toprule
\textbf{} & {Architecture} & \textbf{} & {Architecture} \\\hline
\textit{$G_i$}  & \begin{tabular}[l]{@{}l@{}}fc1($d_t$,512)-relu\\fc2(512,512)-relu\\ fc3(512,512)-relu\\fc4(512,$d_t$)-sigmoid\end{tabular} &
\textit{$De_s$} & \begin{tabular}[l]{@{}l@{}}fc1(1024,2048)-relu\\fc2(2048,$d_s$)\end{tabular} \\ \hline
\textit{$G_s$}  & \begin{tabular}[l]{@{}l@{}}fc1($d_s$,2048)-relu\\fc2(2048,1024)\end{tabular} &
\textit{$De_t$} & \begin{tabular}[l]{@{}l@{}}fc1(1024,2048)-relu\\fc2(2048,$d_t$)\end{tabular}\\ \hline
\textit{$G_t$}  & \begin{tabular}[l]{@{}l@{}}fc1($d_t$,2048)-relu\\fc2(2048,1024)\end{tabular} &
\textit{$D$}     & \begin{tabular}[l]{@{}l@{}}fc1(256,512)-relu\\fc2(512,1)-sigmoid\end{tabular}\\ \hline
\textit{$G$}     & \begin{tabular}[l]{@{}l@{}}fc1(1024,512)-relu\\fc2(512,256)\end{tabular} &
\textit{$F$}     & fc1(256,$n_c$)-softmax \\
\bottomrule         
\end{tabular}}
\end{center}
\end{table}

\begin{table*}[h!]
\begin{center}
\caption{Comparison results over domain adaptation methods on the six constructed digit recognition domain adaptation tasks
	under the setting of 40\% data missing rate and 10 labeled instances per class in the target domain.}
\setlength{\tabcolsep}{2mm}{
\begin{tabular}{c| c c c c c c}
\toprule
Methods &\tabincell{c}{{MNIST}\(\to\)\\{MNIST-M}} & \tabincell{c}{{MNIST-M}\(\to\)\\{MNIST}} & \tabincell{c}{{MNIST}\(\to\)\\{USPS}} & \tabincell{c}{{USPS}\(\to\)\\{MNIST}} & \tabincell{c}{{SVHN}\(\to\)\\{SYN}} & \tabincell{c}{{SYN}\(\to\)\\{SVHN}} \\ \hline
Target only & 0.161\(\pm\)0.003 &	0.592\(\pm\)0.007 &	0.656\(\pm\)0.005 &	0.589\(\pm\)0.005 &	0.117\(\pm\)0.002 &	0.106\(\pm\)0.002\\
DANN & 0.171\(\pm\)0.006 &	0.607\(\pm\)0.006 &	0.669\(\pm\)0.004 &	0.594\(\pm\)0.007 &	0.120\(\pm\)0.003 &	0.159\(\pm\)0.006\\
CDAN & 0.176\(\pm\)0.006 &	0.612\(\pm\)0.006 &	0.673\(\pm\)0.003 &	0.613\(\pm\)0.005 &	0.124\(\pm\)0.003 &	0.167\(\pm\)0.003\\
\hline
IDIAN & \textbf{0.213}\(\pm\)0.004 &	\textbf{0.743}\(\pm\)0.006 &	\textbf{0.775}\(\pm\)0.005 &	\textbf{0.723}\(\pm\)0.006 &	\textbf{0.128}\(\pm\)0.005 &	\textbf{0.184}\(\pm\)0.005\\

\bottomrule
\end{tabular}}
\label{tab:digit10}
\end{center}
\end{table*}

\begin{table*}[]
\begin{center}
\caption{Comparison results over domain adaptation methods on the six constructed digit recognition domain adaptation tasks
	under the setting of 40\% data missing rate and 20 labeled instances per class in the target domain.}
\setlength{\tabcolsep}{2mm}{
\begin{tabular}{c| c c c c c c}
\toprule
Methods &\tabincell{c}{{MNIST}\(\to\)\\{MNIST-M}} & \tabincell{c}{{MNIST-M}\(\to\)\\{MNIST}} & \tabincell{c}{{MNIST}\(\to\)\\{USPS}} & \tabincell{c}{{USPS}\(\to\)\\{MNIST}} & \tabincell{c}{{SVHN}\(\to\)\\{SYN}} & \tabincell{c}{{SYN}\(\to\)\\{SVHN}} \\ \hline
target only & 0.172\(\pm\)0.003 & 0.691\(\pm\)0.004 & 0.750\(\pm\)0.003 & 0.689\(\pm\)0.004 & 0.130\(\pm\)0.004 & 0.113\(\pm\)0.002 \\
DANN & 0.177\(\pm\)0.009 & 0.710\(\pm\)0.007 & 0.753\(\pm\)0.017 & 0.706\(\pm\)0.003 & 0.135\(\pm\)0.008 & 0.192\(\pm\)0.018 \\
CDAN & 0.183\(\pm\)0.004 & 0.714\(\pm\)0.003 & 0.756\(\pm\)0.008 & 0.709\(\pm\)0.008 & 0.137\(\pm\)0.004 & 0.191\(\pm\)0.004 \\
 \hline
IDIAN & \textbf{0.223}\(\pm\)0.002 & \textbf{0.777}\(\pm\)0.004 & \textbf{0.820}\(\pm\)0.010 & \textbf{0.780}\(\pm\)0.006 & \textbf{0.143}\(\pm\)0.004 & \textbf{0.195}\(\pm\)0.023 \\
\bottomrule
\end{tabular}}
\label{tab:digit20}
\end{center}
\vskip -.1in
\end{table*}


\begin{table*}
\begin{center}
\caption{Comparison results over domain adaptation methods on the real world ride-hailing dataset.}
\begin{tabular}{c|ccccc}
\toprule
 Methods & AUC   & ACC   & Recall   & Precision   & F1 score \\ \midrule
Target only  & 0.564\(\pm\)0.006 &	0.564\(\pm\)0.006 &	0.568\(\pm\)0.006 &	0.563\(\pm\)0.006 &	0.566\(\pm\)0.009 \\
DANN  & 0.580\(\pm\)0.002 &	0.580\(\pm\)0.002 &	0.588\(\pm\)0.005 &	0.578\(\pm\)0.006 &	0.584\(\pm\)0.008 \\
CDAN   & 0.585\(\pm\)0.004 &	0.585\(\pm\)0.004 &	0.565\(\pm\)0.009 &	0.582\(\pm\)0.008 &	0.583\(\pm\)0.007 \\
 \hline
	IDIAN & {\bf 0.595}\(\pm\)0.004 & {\bf 0.595}\(\pm\)0.004 & {\bf 0.611}\(\pm\)0.009 &{\bf 0.597}\(\pm\)0.007 &	{\bf 0.604}\(\pm\)0.009 \\  \bottomrule
\end{tabular}
\label{tab:didi1000}
\end{center}
\end{table*}

\begin{table*}
\begin{center}
\caption{The ablation study results on the task MNIST-M$\rightarrow$MNIST.}
\begin{tabular}{lccccc}
\toprule
Methods & 20\% missing   & 40\% missing   & 60\% missing   & 80\% missing \\ \midrule
IDIAN w/o imputation & 0.794\(\pm\)0.006 & 0.664\(\pm\)0.005 & 0.498\(\pm\)0.007 & 0.265\(\pm\)0.006\\
IDIAN w/o $L_{AE}$ & 0.790\(\pm\)0.006 & 0.640\(\pm\)0.008 & 0.453\(\pm\)0.005 & 0.264\(\pm\)0.003\\
IDIAN w/o $L_{contrast}$ & 0.802\(\pm\)0.005 & 0.731\(\pm\)0.006 & 0.561\(\pm\)0.007 & 0.296\(\pm\)0.007\\
\hline
IDIAN & \textbf{0.814}\(\pm\)0.005 & \textbf{0.743}\(\pm\)0.006 & \textbf{0.575}\(\pm\)0.006 & \textbf{0.312}\(\pm\)0.005\\
 \bottomrule
\end{tabular}
\end{center}
\end{table*}
\label{tab:ablation-digit}

\section{Experiment}

We conducted experiments on both benchmark digit recognition datasets for domain adaptation with simulated incomplete target domains 
and a real world 
domain adaptation problem with natural incomplete target domains
for ride-hailing service request prediction. 
In this section, we present our experimental setting and results. 

\subsection{Experimental Settings}
\paragraph{Digit Recognition Image Datasets} 
We used a set of commonly used domain adaptation tasks constructed on five types of digit recognition datasets. 
The five digital datasets are 
MNIST~\cite{lecun1998gradient}, MNIST-M, 
Street View House Numbers (SVHN)~\cite{netzer2011reading}, 
Synthetic Numbers (SYN)~\cite{moiseev2013evaluation} and USPS~\citep{hull1994database}. 
We contructed six common domain adaptation tasks 
by using these datasets as three pairs of domains: 
(1) {MNIST \(\leftrightarrow\) MNIST-M.} 
MNIST-M is obtained from MNIST by 
blending digits from the original set over patches randomly extracted from color photos from BSDS500.
We can have two domain adapation tasks by using each one as the source domain and the other one as the target domain.
(2) {SYN \(\leftrightarrow\) SVHN}. 
Synthetic numbers (SYN) consists of 500,000 synthesized images generated from Windows fonts. 
We put this {\em synthesized} digit image set together with 
{\em real} Street-View House Number dataset (SVHN) as adaptation domain pairs. 
Again, two domain adaptation tasks can be obtained by using one domain as the source domain and 
the other domain as the target domain, and then reversing the order.  
(3) {MNIST \(\leftrightarrow\) USPS.} 
In the same manner as above, we also constructed two domain adaptation tasks 
between the USPS handwritten digit images and the MNIST set. 
We used an unsupervised Autoencoder model to extract features from raw images on each dataset,
which we later used as the input data in our domain adaptation experiments. 
The encoder of the model consists of three convolutional layers, 
while the decoder is composed of three transpose convolutional layers. 
We resize each image to 32 x 32 x 3 as the input of the autoencoder,
and the encoder maps each image into a 1024-dim feature vector.

As these standard domain adaptation tasks have fully observed data in both domains,
we simultate the incomplete target domain 
by randomly setting part of the instance feature values as zeros in the target domain, 
indicating the missing status of the corresponding entries. 
We can create incomplete target domains with any feature missing rate between 0 and 1.  
Moreover, to further enhance the difference of the cross-domain features,
we also randomly shifted the order of the feature channels in the target domains. 

\paragraph{Ride-Hailing Service Request Adaptation Dataset}
We collected a real world adaptation dataset with incomplete target domains from 
a ride-hailing service platform.
The advertising needs on the ride-hailing service platform often requires
the prediction of the service usage of new users given the historical service usage data of many active users.
We treat this problem as a cross-domain binary classification problem over users,
where the active users' data form the source domain and the new users' data form the target domain. 
As a new user's information typically contains many missing entries,
the target domain in this problem is naturally incomplete. 
We obtained a source domain with 400k instances of active users 
and a target domain with 400k instances of new users. 
Moreover, as the active users and the new users are collected in different time and manner, 
there is no record of the feature space correspondence between them 
though they do share many attributes. 
In the dataset, the feature dimension in the source domain is 2433 and 
in the target domain is 1302. 
Moreover, the feature missing rate in the target domain is very high, close to 89\%.

\paragraph{Model Architecture}
For the proposed IDIAN model, we used the multi-layer perceptrons for its components.
Specifically, we used 
a four layer network
for $G_i$.
The feature extractors $\{G_s, G_t, G\}$, the decoders $\{De_s, De_t\}$, 
and the discriminator $D$ are each composed of two fully connected layers respectively. 
The classifier $F$ is composed of one fully connected layer.
The specific details are provided in Table~\ref{tab:modelarch}.


\paragraph{Comparison Methods}
This is the very first work that addresses the problem of domain adaptation with incomplete target domains.
Moreover, our problem setting is very challenging such that the input feature spaces of the two domains can be different, 
Hence we compared our proposed IDIAN model with the following baseline and two 
adapted state-of-the-art adversarial domain adaptation methods:
(1) {\em Target only}. This is a baseline method without domain adaptation, 
which trains a classification network with only the labeled data in the target domain. 
For fair comparison, we used the same architectures of feature extractor ($G_t$ and $G$) 
and classifier ($F$) as our proposed model.
(2) {\em DANN}. This is an adversarial domain adaptation neural network developed in~\cite{ganin2016domain}.
For fair comparison and also adapting DANN to handle different cross-domain feature spaces,
we build DANN under the same framework as our proposed model
by dropping $G_i$, $De_s$ and $De_t$, while only using the adversarial loss and classification loss
as the optimization objective. 
(3) {\em CDAN}. This is a conditional adversarial domain adaptation network developed in~\cite{long2018conditional}.
It takes the instance's class information as a joint input to the adversarial domain discriminator,
aiming to address the multimodal structure of the feature alignment. 
Here, we build CDAN by adjusting the DANN above and providing the classifier's label prediction results
as input to the conditional adversarial domain discriminator. 

\subsection{Experiments on Image Datasets}   

For each of the six domain adaptation tasks constructed on the digital image datasets,
we simulated the incomplete target domain in different situations by dropping out 
20{\%}/40{\%}/60{\%}/80{\%} of the feature values respectively.
We also conducted experiments by randomly selecting 10 or 20 labeled instances from each category 
as the labeled instances in the target domain and using the rest target data as unlabeled data.

In this set of experiments, we used a learning rate $\eta= 0.01$ and set the batch size to 128. 
The trade off parameters of IDIAN (\({\alpha, \beta, \gamma, \lambda}\)) are set 
as (1,10,10,10). We set the epoch number $N_e$ as 20.
We repeated each experiment five times,
and recorded the mean accuracy and standard deviation values of the results on the test data of the target domain.

\paragraph{Results} 
Table~\ref{tab:digit10} and Table~\ref{tab:digit20} report the comparison results 
on the six domain adaptation tasks with a 40\% feature missing rate in the target domain
by using 10 and 20 instances from each class in the target domain as labeled instances respectively. 
We can see that in both tables, the {\em Target only} baseline produces the worst results
across all domain adaptation tasks. 
With domain adaptation, both {\em DANN} and {\em CDAN}
outperform {\em Target only} with notable margins,
while {\em CDAN} produces even slightly better results than {\em DANN}.
Nevertheless, the proposed IDIAN produced the best results 
among all the comparison methods across all the six tasks. 

We also experimented with different feature missing rates in the target domain. 
The six sub-figures in Figure~\ref{fig:digit10} present the comparison results on 
the six domain adaptation tasks respectively across multiple feature missing rates (20\%, 40\%, 60\%, 80\%)
in the target domain. 
Again, we can see our proposed IDIAN consistently outperforms all the other methods across all scenarios. 
These results demonstrated the efficacy of our proposed model.

\subsection{Experiments on Ride-Hailing Dataset}
On this real world incomplete domain adaptation task, 
we used 50\% of the target domain data for training and the remaining 50\% for testing. 
On the training data, we randomly chose 1000 instances in the target domain as 
labeled instances ($n_l=1000$). All the data in the source domain are used as training data.
We used a learning rate $\eta=0.01$ and set the batch size as 500. 
We set the trade off parameters (\({\alpha, \beta, \gamma, \lambda}\)) as (5,20,20,20)
respectively, and set the epoch number as 50.
We repeated the experiment five times,
and recorded the mean and standard deviation values of the test results. 

\paragraph{Results} 
For this binary classification task, we evaluated the test performance 
using five different measures: AUC, ACC (accuracy), recall, precision and F1 score.
The comparison results are reported in Table~\ref{tab:didi1000}. 
We can see that, similar to previous results, 
all the domain adaptation methods outperform the {\em Target only} baseline.
This verified the efficacy of the domain adapation mechanism even in this
much challenging real world learning scenario. 
Moreover, the proposed IDIAN further outperforms both DANN and CDAN 
in terms of all the five different measures. 
In terms of F1 score, IDIAN outperforms the baseline by 3.8\%.
The results validated the efficacy of our proposed model.

\subsection{Ablation Study}
To further analyze the proposed IDIAN model,
we conducted an ablation study 
on the adaptation task from MNIST-M$\rightarrow$ MNIST with 
10 labeled instances from each target class.
Specifically, we compared the full IDIAN model with the following three variants:
(1) IDIAN w/o imputation. This variant drops the incomplete data imputation component in IDIAN. 
(2) IDIAN w/o $L_{AE}$. This variant drops the decoders and the Autoencoder loss $L_{AE}$ in IDIAN. 
(3) IDIAN w/o $L_{cont}$. This variant drops the inter-domain contrastive loss $L_{cont}$ in IDIAN.
The comparison results are reported in Table~\ref{tab:ablation-digit}
We can see that the performance of all the variants are much inferior to the full IDIAN model.
The results are consistent across settings with different target feature missing rates,
which validated the essential contribution of the data imputation, autoencoder,
and inter-domain contrastive loss for the proposed IDIAN model.


\section{Conclusion}
 
In this paper, we addressed a novel domain adaptation scenario where the data in the target domain are incomplete.
We proposed an Incomplete Data Imputation based Adversarial Network (IDIAN) model to 
address this new domain adaptation challenge.
The model is designed to handle both homogeneous and heterogeneous cross-domain feature spaces. 
It integrates data dependent feature imputation, autoencoder-based cross-domain feature space unification,
and adversarial domain adaptation coherently into an end-to-end deep learning model. 
We conducted experiments on both cross-domain benchmark tasks with simulated incomplete target domains 
and a real-world adaptation problem on ride-hailing service request prediction with natural incomplete target domains.
The experimental results demonstrated the effectiveness of the proposed model. 

\bibliography{paperbib}

\end{document}